\pdfoutput=1
\documentclass{article} % For LaTeX2e
\usepackage{nips11submit_e,times}

\usepackage{graphicx} % more modern
\usepackage{subfigure} 
\usepackage{amsmath,amsthm}

\usepackage{algorithm}
\usepackage{algorithmic}
\usepackage{fancyvrb}
\newcommand{\namelong}{Linear Denoiser}
\newcommand{\name}{LIDE} 
\newcommand{\snamelong}{Stacked Linear Denoiser}
\newcommand{\sname}{SLIDE} 
\def\tm{\leavevmode\hbox{$\rm {}^{TM}$}}

\title{Rapid Feature Learning with Stacked Linear Denoisers }

\author{
Zhixiang Eddie Xu\\
Department of Computer Science\\
Washington University in St. Louis\\
Saint Louis, MO 63108 \\
\texttt{xuzx@cse.wustl.edu} \\
\And
Kilian Q. Weinberger \\
Department of Computer Science \\
Washington University in St. Louis\\
Saint Louis, MO 63108 \\
\texttt{kilian@wustl.edu} \\
\AND
Fei Sha\\
Computer Science Department \\
University of Southern California \\
Los Angeles, CA 90089\\
\texttt{feisha@usc.edu} \\
}

\begin{document} 

\nipsfinalcopy
\maketitle
% \nipsfinalcopy

\newcommand{\argmax}{\operatornamewithlimits{argmax}}
\newcommand{\argmin}{\operatornamewithlimits{argmin}}
\newcommand{\x}{\mathbf{x}}
\newcommand{\q}{\mathbf{q}}
\newcommand{\bb}{\mathbf{b}}
\newcommand{\h}{\mathbf{h}}
\newcommand{\ab}{\mathbf{\alpha}}
\newcommand{\y}{\mathbf{y}}
\newcommand{\Lb}{\mathbf{L}}
\newcommand{\Wb}{\mathbf{W}}
\newcommand{\Ib}{\mathbf{I}}
\newcommand{\Kb}{\mathbf{K}}
\newcommand{\Hb}{\mathbf{H}}
\newcommand{\Qb}{\mathbf{Q}}
\newcommand{\Pb}{\mathbf{P}}
\newcommand{\Xb}{\mathbf{X}}
\newcommand{\Sb}{\mathbf{S}}

\newcommand{\fix}{\marginpar{FIX}}
\newcommand{\new}{\marginpar{NEW}}

\begin{abstract}
We investigate unsupervised pre-training of deep architectures as feature generators for ``shallow'' classifiers. 
%Inspired by an in-depth study of why pre-training improves the results of deep neural networks~\cite{erhan2010does}, 
Stacked Denoising Autoencoders (SdA)~\cite{Vincent08extractingand}, when used as feature pre-processing tools for SVM classification, can lead to significant improvements in accuracy -- however, at the price of a substantial increase in computational cost. 
In this paper we create a simple algorithm which mimics the layer by layer training of SdAs. However, in contrast to SdAs, our algorithm  requires no training through gradient descent as the parameters can be computed in closed-form. It can be implemented in less than 20 lines of MATLAB\tm  and reduces the computation time from several hours to mere seconds. We show that our feature transformation reliably improves the results of SVM classification significantly on all our data sets -- often outperforming SdAs and even deep neural networks in three out of four deep learning benchmarks. 
%we investigate the hypothesis that pre-training can be viewed as a feature generating process. We explore supervised and semi-supervised settings, where we pre-train a denoising autoencoder and use the values of the hidden representations as input to a SVM classifier. We demonstrate empirically, on several difficult benchmark data sets, that this unsupervised pre-training step can significantly improve the accuracy of the SVM predictor --- often outperforming fine-tuned deep neural networks that are initialized with the weights of the autoencoder. 
% This trend is also prominent in semi-supervised settings, where the pre-training is performed on additional unlabeled data. 
\end{abstract}

\section{Introduction}
\label{sec_introduction}
Recently, there has been a great deal of attention on ``deep-learning'' architectures~\cite{bengio2007scaling,hinton2006fast}. Such architectures have consistently achieved state-of-the-art results on many challenging learning tasks, including object recognition~\cite{nair20093}, natural language processing~\cite{collobert2008unified},  dimensionality reduction~\cite{hinton2006fast}. A typical paradigm of deep-learning is to first perform unsupervised pre-training on the neural network to initialize the weights and then 
use back-propagation for supervised training.
%train a multi-layer neural network with back initialize the parameters through an unsupervised pre-training step. %Multiple approaches have been proposed for pre-training, including restricted Boltzmann machines~\cite{Hinton07282006} and de-noising autoencoders~\cite{larochelle2009exploring}. 
%
%A big component of deep-learning algorithms is the unsupervised pre-training which reduces the main shortcomings of neural networks -- in particular the chance of getting stuck in bad local minima.  

%Although the features obtained with SdA can lead to great accuracy improvements in many learning tasks~\cite{lee2009unsupervised}, the overhead of training and model-selection for SdA is substantial. 

\cite{lee2009unsupervised} showed that the second phase, the supervised back-propagation, can be replaced with ``shallow'' classifiers, such as support vector machines (SVM)~\cite{cortes1995support}. Concretely, the outputs of the hidden units of pre-trained deep neural networks are  used as input features to those classifiers. Besides achieving superior performance in recognition tasks,  the substituting classifiers offer appealing computational properties. In particular, their parameters are adjusted with convex optimization, free of local optima that often plague back-propagation based techniques. Further, these classifiers tend to be more ready to be used out-of-the-box and can often be parallelized very effectively.  
% and are often more parallelizeable and potentially 
%Additionally, these classifiers can be implemented in parallel, thus are potentially much 
%more scalable than the sequential paradigm of stochastic gradient descent used in back-propagation.

%In this learning setup, the pre-training becomes a process to generate deep features~\cite{Vincent08extractingand}. 
\emph{How can we replace the pre-training phase with an equally attractive learning model}?  Note that the  pre-training phrase -- now merely used for unsupervised feature generation -- is a major bottleneck in applying deep learning architectures. One needs to adjust several parameters of network architectures (the number of hidden layers, units in each hidden layer), optimization (learning rates and momentum), etc. Compounded by  these factors, the pre-training phase takes up a large portion of the overall training time  (anecdotally 50-90\%) -- even with the use of multi-core processors and graphical processing units (GPUs)~\cite{bengio2007speeding}. %Thus, alternative approaches that are  computationally efficient  are highly desirable and have broad application potentials in 
%applying deep learning architectures to extremely large-scale problems.

We propose such a method for pre-training, thus for feature generation, and investigate its effectiveness in this paper.  We show how one-layer \emph{linear} denoising autoencoders can be used as the basic building blocks of the pre-training phases. The autoencoders are ordinary linear regression models for reconstructing data from corrupted features. They have closed-form solutions and thus are easy to implement. In particular, the parameters can be identified with matrix inversion and nonlinear optimization is not needed. We use the outputs of linear denoising autoencoders in two ways: i) as input features to support vector machines; ii) as inputs (to be denoised) to successively stacked linear autoencoders.  

The proposed method, which we refer to as \emph{\snamelong{}} (\sname{}),  is similar in spirit to stacked (nonlinear) denoising autoencoders
 (SdA)~\cite{Vincent08extractingand}. However,  there are important differences. 
First, in SdA, hidden layers are used to extract new representations from inputs which makes their training a nonconvex optimization. In contrast, \sname{} does not have hidden layers and is convex. Second,  SdA  requires the setting of several crucial meta-parameters, including noise level, learning rates, number of training epochs and network architecture specifics, which are typically set by cross-validation.  In comparison, \sname{} only has \emph{two} free meta-parameters, controlling the amount of noise to be added to data as well as the number of autoencoders we would like to stack. Finally, leveraging on the analytic tractability of linear regression, we train the parameters of our autoencoders to optimally denoise \emph{all possible} corrupted training inputs --- arguably ``\emph{infinitely many}''. This is practically feasible for SdA, whose parameters are adjusted on only a subset of corrupted data.
 
In this paper, we make several contributions.  \sname{} can be implemented easily (less than 20 lines of Matlab codes). The learning algorithm runs very fast. In fact, our 19-lines implementation is  \emph{three orders of magnitudes} faster than a highly optimized parallel implementation of SdA on a state-of-the-art GPU.  Even for large data sets with tens of thousands of samples the computation takes only a few seconds -- achieving 
%On standard benchmark datasets for deep learning, we achieve 
thousand-fold speed-up over the pre-training phase with SdA or  similar approaches. 
%When we use \sname{} features as input to SVMs we obtain
%The autoencoders' outputs are treated as input features to support vector machines. We also obtain superior classification results on several deep learning benchmarks: In 3 out of 4 tasks, we obtain much improved results than standard deep learning neural networks.  
%In the experimental section 
Finally, in addition to the vastly reduced computation time, we demonstrate on several deep-learning benchmark data sets that  SVM classification  with \sname{} features tends to be even more accurate than with SdA features or pre-trained deep neural networks with back-propagation.

\section{Notation and Background}
\label{sec_setting}
In the following, we introduce notation and algorithms which are used in the rest of the paper. Our training data consist of $n$ input vectors $\left\{\x_1,\dots,\x_n\right\}\in{\cal R}^d$  with corresponding discrete class labels $\left\{y_1,\dots,y_n\right\}$ drawn from an unknown joint distribution ${\cal D}$. 
%Our goal is to learn a feature transformation $\x_i\rightarrow f(\x_i)$ to improve the generalization error $E_{(\x,y)\sim {\cal D}}\left[h(f(\x))\neq y\right]$  of a classifier $h(\cdot)$, trained to minimize the loss $\frac{1}{n}\sum_{i=1}^n \left[h(f(\x_i))\neq y_i \right]$. 
%Given a hypothesis class ${\cal H}$

% For simplicity, we assume that the vectors $\x_i$ are appropriately scaled such that $\x_i\in [0,1]^d$. 
%Our goal is to learn a good feature representation 
%after the pre-training, 
%the predictor $h:{\cal R}^d\rightarrow {\cal Y}$ can attain the empirical minimum risk, that is, $h(\x_i)$ is as close to $y_i$ as possible (measured by some loss function ${\cal L}(h)$). 

\textbf{Deep Architecture.}
Deep learning algorithms learn hierarchies of hidden layers, where the output of the lower level layer becomes the input of the higher level. %The term ``deep learning'' was coined by~\cite{Hinton07282006}, and started a renaissance of the traditional neural network as a popular learning algorithm.
``Deep learning'' algorithms differ from traditional neural networks in two ways: i) they tend to have more hidden layers (\emph{i.e.} are \emph{deeper}); ii) the supervised training through back-propagation is preceded by an unsupervised pre-training step in which the weights are initialized in a generative manner. 
The additional layers are believed to provide more powerful learning models. The pre-training makes the training more efficient (it is often performed greedily -- layer by layer) and regularizes the optimization so that back-propagation starts near a ``good'' local minima~\cite{erhan2010does}. %Recently, it has been shown~\cite{lee2009unsupervised} that pre-trained deep neural networks, when used for unsupervised feature pre-processing, can significantly improve the generalization error of SVM classifiers. 

%In contrast to previous approaches, where training was performed only through back-propagation~\cite{rumelhart2002learning} -- starting from randomly initialized weights, the ``deep learning'' by Hinton \emph{et al.} adds a preceding step of unsupervised pre-training of the weight vectors. This is particularly powerful when the neural network has multiple hidden layers. 

%However, for the deep neural networks, the gradient descend optimization and random initialization employed by the traditional neural networks may suffer from very poor local minimal. For the deep belief nets, the explaining away in the directed graphic model makes the deep nets much more difficult to train.
%Recently, the situation has been changed by some approaches from ~\cite{Hinton06afast}~\cite{Vincent08extractingand}~\cite{Bengio_1learning}, which provide efficient ways to train a deep architecture.

\textbf{Support Vector Machines} are one of the most popular and reliable out-of-the-box supervised classification algorithms. 
SVMs~\cite{Scholkopf02} are linear classifiers that involve a quadratic minimization problem, which is convex and not plagued by local optima. The maximum margin separation promotes reliably good generalization, and the \emph{kernel-trick}~\cite{Scholkopf02} allows SVMs to generate highly non-linear decisions boundaries with low computational overhead. 
The kernel-trick maps the input vectors $\x_i$ implicitly into a higher (possibly infinite) dimensional feature space using the kernel function $k(\x_i,\x_j)$. Among various such functions, the Radial Basis Function (RBF)-Kernel, is one of the most commonly used kernels. When used with Euclidean distances (also typically referred to as the Gaussian kernel) it is defined as follows:
\begin{equation}
k(\x_i,\x_j) =\exp\left({-\frac{\|\x_i-\x_j\|^2_2}{\sigma^2}}\right),\label{eq:rbf}
\end{equation}
%where $d(\cdot,\cdot)$ denotes the Euclidean distance between inputs $\x_i$ and $\x_j$. 
where $\sigma^2$ denotes the RBF-\emph{kernel-width}. 
%The two SVM parameters, the global kernel-width $\sigma$ and the regularization constant $C$ (for more details see~\cite{cortes1995support}), are usually picked by cross-validation on a validation split.

\textbf{Stacked Denoising Autoencoder.}
One particular method to pre-train the weights of a deep neural network is Stacked Denoising Autoencoders (SdA)~\cite{Vincent08extractingand,larochelle2009exploring}. A traditional autoencoder described in~\cite{Hinton07282006} maps the input data, or visible features $\left\{\x_1,\dots,\x_n\right\}\in{\cal R}^d$ into hidden representations $\left\{\h_1,\dots,\h_n\right\}\in{\cal R}^m$. The hidden representations are then mapped back to reconstructions of the original input $\left\{\x^{\prime}_1,\dots,\x^{\prime}_n\right\}\in{\cal R}^d$. For both mappings they use the sigmoid function
\begin{equation}
	\h_i=\frac{1}{1+e^{-(\Wb\x_i+\bb)}} \ \ \textrm{ and }\ \   	 {\x}_i'=\frac{1}{1+e^{-(\Wb^\top\h_i+\bb')}},\label{eq:da_sigmoid}
\end{equation}
where the parameters consist of a weight matrix $\Wb\in{\cal R}^{m\times d}$ and two bias vectors $\bb,\bb'\in{\cal R}^{m}$.  %using the same sigmoid function as~(\ref{eq:sigmoid}), but with different parameters $\Wb^{\prime}$ and $b^{\prime}$. 
% Because we use the transposed weight matrix $\Wb^\top$ for reconstruction, the autoencoder is said to have \emph{tied weights}. % when $\Wb^{\prime} = \Wb^\top$. 
The optimal parameters are learned by minimizing the reconstruction error, which is measured in the squared loss
\begin{equation} 
{\cal L}_{sq}(\Wb,\bb,\bb^{\prime}) = \frac{1}{2}\displaystyle\sum_{i=1}^n\|\x_i - \x^{\prime}_i\|^2. \label{eq:da_sq_loss}
\end{equation}
% Alternative loss functions, for example cross-entropy, could also be used. For simplicity, throughout this work we will focus on the squared-loss only. 
% or the cross-entropy that evaluates the similarity of $\x$ and $\x^{\prime}$, 
% \begin{equation} 
% {\cal L}_{en}(\Wb,b,\Wb^{\prime},b^{\prime}) = \displaystyle\sum_{i=1}^n \x_i \log (\x^{\prime}_i) + (1-\x_i) \log (1-\x^{\prime}_i). \label{eq:entropy_loss}
% \end{equation}

The denoising autoencoder (DA) incorporates a slight modification from the traditional autoencoders.  %It  artificially corrupts each input and reconstructs the original uncorrupted version. More explicitly, f
For each input $\x_i$, it picks a fixed percentage of the features uniformly at random and sets them to zero (effectively removing the features), while keeping others untouched. 
%by uniformly sampling a fixe certain percentage of input data by uniformlly  deleting some entries of the input, or specifically, setting some entries to zero. 
It then maps this corrupted input $\tilde{\x}_i$ into the hidden representations $\tilde{\h}_i$, which is then mapped back as $\tilde{\x}_i'$ to reconstruct the original uncorrupted input, minimizing  %$\left\{\tilde{\x}^{\prime}_1,\dots,\tilde{\x}^{\prime}_n\right\}\in{\cal R}^d$. Training a denoising autoencoder is slightly different from training a traditional one, as the goal is to minimizing the reconstruction error of the corrupted data and filling in those deleted information.
\begin{equation} 
{\cal L}_{sq}(\Wb,\bb,\bb^{\prime}) = \frac{1}{2}\displaystyle\sum_{i=1}^n\|\x_i - \tilde{\x}^{\prime}_i\|^2. \label{eq:sq_loss2}
\end{equation}

%For example, in the case of images, a pixel can usually be well re-constructed from its neighboring pixels. To reduce the impact of noise, a learning algorithm should leverage these known feature correlations and convolute their values. 

The stacked denoising autoencoder~\cite{Vincent08extractingand} stacks several DAs together by feeding the hidden representations of the $i^{th}$ DA as input into the $(i+1)^{th}$ DA. The training is performed greedily layer by layer: When the $(i+1)^{th}$ layer is trained, all layers $1...i$ are fixed and noise is only added to the hidden nodes of the $i^{th}$ layer. 
%This forward feeding structure enforces that the $(i+1)^{th}$ autoencoder is always trained after the $i$-th one, and therefore the whole training is done layer-wised in a greedy way. 

Intuitively, by forcing removed features to be reconstructed from the remaining data, the DA learns to convolute features that tend to be correlated. This increases robustness against noise and local transformations, e.g. small translation or rotation. 
A similar approach has been successfully used for many years in Convolutional Neural Networks (CNN)~\cite{lecun1995convolutional}, which leverage the fact that in natural images local pixels are highly correlated with each other --- which is hard-coded into the network structure.  DAs are more general as they \emph{learn} the convolution patterns, and can be applied to data sets where the feature correlation is unknown (for example in contrast to CNNs, DAs are invariant to arbitrary permutations of the input pixels). However, SdAs suffer from two inherent down-sides: their long training time and sensitivity to several hyper-parameters such as network architecture, learning rates, etc. Carefully tuning these parameters on validation datasets is often very time consuming.

\textbf{Pre-training as Feature Generator}. Many researchers have noticed that the pre-training phase in deep learning networks can be seen as some kind of nonlinear feature mappings. For example, \cite{lee2009unsupervised}  showed that  the hidden representations computed by either all or partial layers of stacked denoising autoencoders make excellent   features for classification with support vector machines. \cite{cho2010large} introduced recursively defined kernels which mimic the pre-training of deep feature extractors for the use with support vector machines. Our work follows this line of thinking and shows simpler and more computationally tractable models can also be used for similar purposes.

%As has been shown by~\cite{Bengio07greedylayer-wise} and~\cite{lee2009unsupervised} de-noising auto-encoders are very effective feature generators for neural networks and support vector machines. 

\section{Single Layer Construction}
\label{sec_method}
Instead of using an autoencoder for pre-training, we propose to reconstruct randomly corrupted data with a linear mapping $\Wb:{\cal R}^d\rightarrow {\cal R}^d$. To lower the variance, we perform multiple passes over the training set, essentially corrupting  $m$ copies of the original data, and solving for the $\Wb$ that minimizes the overall squared loss
%We learn this mapping through minimization of the squared reconstruction error 
% \begin{equation} 
% {\cal L}_{sq}(\Wb,b) = \frac{1}{2}\displaystyle\sum_{i=1}^n\|\x_i - \Wb\tilde{\x}_i\|^2, \label{eq:sq_loss_wb}
% \end{equation}
\begin{equation} 
{\cal L}_{sq}(\Wb) = \frac{1}{2}\displaystyle\sum_{j=1}^m\sum_{i=1}^{n}\|\x_{i} - \Wb\tilde{\x}_{ij}\|^2,\label{eq:ols}
\end{equation}
where $\tilde{\x}_{ij}$ represents the $j^{th}$ corrupted version of the original input $\x_i$. An input is corrupted by setting each feature randomly to zero with probability $(1-p)$. 
%
%Where $\x_i$ is the original input, $\tilde{\x}$ is the corrupted input, $\Wb$ is the linear denoising matrix, and $b$ is the bias. 
To simplify notation we assume that a constant feature is added to the input, $\x_i=[\x_i^\top,1]^\top$, and an appropriate bias is incorporated within the mapping $\Wb=[\Wb,\mathbf{1}]$. The constant feature is never corrupted. 
%
% We can also express the square loss in a matrix form, when we have multiple data points.
% \begin{align}
% 	\displaystyle\sum_{i=1}^m\|(\x_{i}-\Wb\tilde{\x}_{i})\|^2 & = (X-\Wb\tilde{X})^2  \label{eq:sq_loss_matrix} \\
% 	 & = tr(X-\Wb\tilde{X})^{\top}(X-\Wb\tilde{X}) \notag
% \end{align}
% 
% Our goal is to find the denoising matrix $\Wb$ that minimizes the square loss~(\ref{eq:sq_loss_matrix})
% \begin{equation}
% 	\Wb = \arg\min_{\Wb}(X-\Wb\tilde{X})^2 \label{eq:argminw}
% \end{equation}
For notational simplicity, let us define the $m$-times repetitions of the design matrices $\Xb=[\x_1,\dots,\x_n]$ as $\bar \Xb=[\Xb,\dots,\Xb]\in{\cal R}^{d\times nm}$. Let $\tilde \Xb$ be the corrupted equivalent of $\bar \Xb$, \emph{i.e.} $\tilde \Xb=[\tilde\x_{1,1},\dots,\tilde\x_{n,1},\tilde\x_{1,2},\dots,\tilde\x_{n,m}]$. We can then define two  (scaled) outer-product matrices as:
\begin{equation}
	\Qb=\frac{1}{m}\tilde \Xb\tilde \Xb^\top \textrm{ and } \Pb=\frac{1}{m}\bar \Xb \tilde \Xb ^\top. \label{eq:pandq}
\end{equation}
Note that only $\mathbf{P}$ is defined over the corrupted and uncorrupted data. 
We can then express eq.~(\ref{eq:ols}) as the well-known closed-form solution for ordinary least squares~\cite{Bishop06}:
\begin{equation}
	\Wb = \Pb \Qb^{-1}.	\label{eq:solution}
\end{equation}
% In case of the singular matrix, we add a small regularizer into the solution.
% \begin{equation}
% 	\Wb = (\tilde{X}^{\top} \tilde{X} + \lambda \Ib)^{-1} \tilde{X}^{\top} X	\label{eq:solution_lamda}
% \end{equation}

\textbf{Virtual Denoising.}
% \section{Virtual Denoising}
% \label{sec:virtual}
\begin{algorithm}[t]
\caption{A single layer construction in MATLAB\tm.\label{code:matlab}}
\begin{center}
%\parbox{0.4\textwidth}{
\begin{Verbatim}[frame=none, framerule=0.1mm]
   function [W,h]=lide(X,p);
    X=[X;ones(1,size(X,2))];
    d=size(X,1);
    q=ones(d,1).*p;
    q(end)=1;
    S=X*X';
    Q=S.*(q*q');
    Q(1:d+1:end)=q.*diag(S);
    P=S.*repmat(q,1,d);
    W=((Q+1e-5*eye(d))\P(:,1:end-1))';
    h=W*X;\end{Verbatim}
\end{center}
%\caption{MATLAB code for \name{}. The input is considered a $d\times n$ matrix $x$, where each column represents one input.\label{code:matlab}}

\end{algorithm}
%
%However, it is difficult to determine how many copies are enough to attain a low variance denoising matrix $\Wb$. 
The larger $m$, the more different corruptions we average over. Ideally we would like $m\!\rightarrow \!\infty$, effectively using infinitely many copies of noisy data to compute the denoising transformation $\Wb$. 
%the $m$ in~(\ref{eq:sq_loss_more}) $m\rightarrow \infty$, the solved denoising matrix $\Wb$ should converge to its expected value. 
%Refer to Bishop (12.56). 

By the weak law of large numbers, the matrices $\Pb$ and $\Qb$, as defined in eq.~(\ref{eq:pandq}), converge to their expected values as $m$ becomes very large. If we are interested in the limit case, where $m\rightarrow \infty$, we can derive a closed-form for these expectations and express the corresponding mapping $\Wb$ as
\begin{equation}
	\Wb=E[\Pb]E[\Qb]^{-1}.\label{eq:Wlimit}
\end{equation}
In the remainder of this section, we compute the expectations of $\Qb$ and $\Pb$. For now let us focus on $\Qb$, whose expectation is defined as
\begin{equation}
	E[\Qb]=\sum_{i=1}^{nm} E\left[\tilde\x_{i}\tilde\x_{i}^\top\right].
\end{equation}
The off-diagonal entries of $\tilde \x_{i}\tilde \x_{i}^\top$ are uncorrupted if the two features both ``survived'' the corruption, which happens with probability $p^2$. For the diagonal entries this holds with probability $p$. Let us define a vector $\q=[p,\dots,p,1]^\top\!\in\!{\cal R}^{d+1}$, where $\q_\alpha$ represents the probability of a feature $\alpha$ ``surviving'' the corruption. As the constant feature is never corrupted, we have $\q_{d+1}=1$.  
If we further define the scatter matrix of the original uncorrupted input as $\Sb=\Xb\Xb^\top$, we can express the expectation of the matrix $Q$ as
\begin{equation}
E[\Qb]_{\alpha,\beta}	 = \left\{  
	\begin{array}{lcc}
		\Sb_{\alpha\beta} \q_\alpha \q_\beta & if & \alpha\neq \beta \\
		\Sb_{\alpha\beta} \q_\alpha   & if & \alpha=\beta \label{eq:expected_w}
   	\end{array}.\right.
\end{equation}
By analogous reasoning, we obtain the expectation of $\Pb$ in closed-form as $E[\Pb]_{\alpha\beta}=\Sb_{\alpha\beta}\q_\beta$. 

We refer to this linear transformation as \namelong{} (\name{}). Algorithm~\ref{code:matlab} shows a 10-lines MATLAB\tm implementation. 
\begin{figure}[t]
	\centerline{	\includegraphics[width=0.75\textwidth]{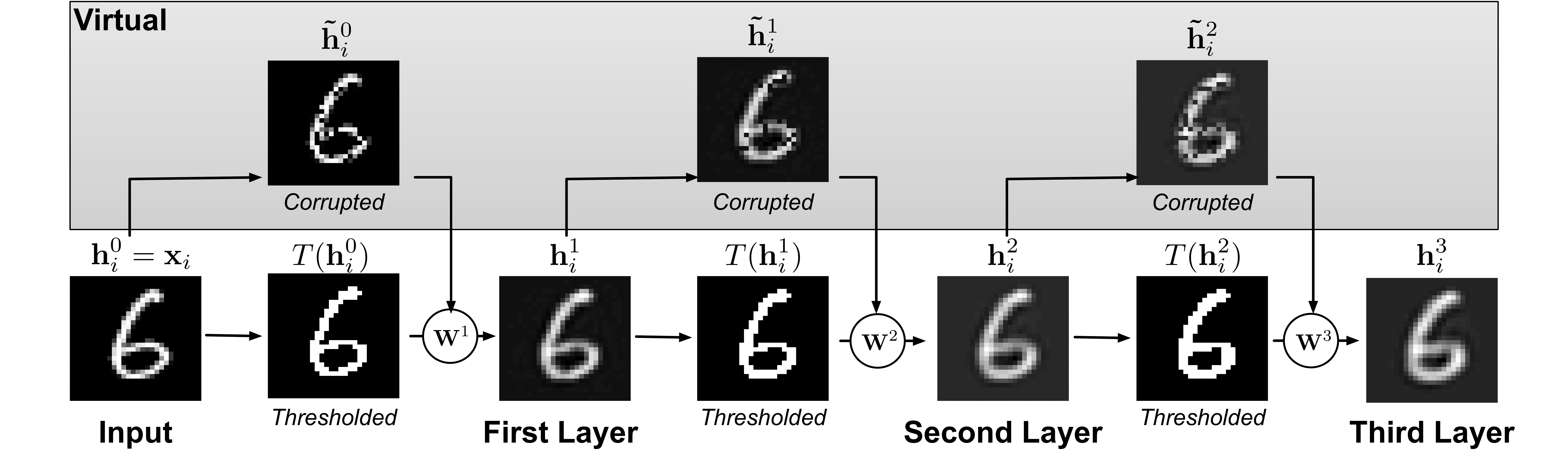}}
	\caption{A schematic description of \sname{}. The Corruption is ``virtual'' in a sense that images are never actually corrupted, as the matrices $\Wb^k$ can be computed directly in closed-form.}
	\label{fig:flow}                          
\end{figure}

\section{Stacked Feature Generation}
\label{sec:stacked}
Some of the success of SdAs can be attributed to the fact that they learn \emph{deep} internal representation. So far, \name{} consists of a linear transformation and therefore cannot compete in terms of feature expressiveness. 
%In order to represent non-linear functions in the real world, we add a non-linear operation into our algorithm, and also stack layers together, increasing the level of non-linearities. 
Inspired by the layer-wise stacking of DAs and DBNs, we stack several \name{} layers together by feeding the representations of the $k^{th}$ denoising layer as the input to the $(k+1)^{th}$ layer. The training is performed greedily layer by layer: When the $(i+1)^{th}$ layer is trained, all layers $1...l$ are fixed, which means we only learn the $(k+1)^{th}$ denoising matrix $\Wb^{k+1}$. For a given input $\x_i$, let $\h_i^k$ denote the output of the $k^{th}$ \name{} transformation. For notational simplicity let us denote  $\h_i^0=\x_i$.

To be able to move beyond a linear transformation, we need to apply a non-linear ``squashing''-function between the layers. 
%The non-linear operation we apply is similar to the Max-Pooling of CNN~\cite{lecun1995convolutional}. 
Several choices might be possible, including sigmoid, or $\tanh$. However in our experiments we simply use a threshold function \mbox{$T(a)=\delta_{a>t}\in\{0,1\}$} for some $t$, which we apply element-wise on vectors.

% To be able to move beyond a linear transformation, we need to apply a non-linear sigmoid function between the layers, similar to SdA. 
%The non-linear operation we apply is similar to the Max-Pooling of CNN~\cite{lecun1995convolutional}. 
% Several choices might be possible, including sigmoid, or $\tanh$. However in our experiments we simply use a threshold function \mbox{$T(a)=\delta_{a>t}\in\{0,1\}$} for some $t$, which we apply element-wise on vectors.

We obtain each layer's representation from the previous layer through the transformation \mbox{$\h^k=\Wb^{k}T(\h^{k-1})$}, \emph{i.e.} we threshold the input  before the denoising transformation. 
Analogously to eq.~(\ref{eq:ols}), each transformation $\Wb^{k+1}$ is learned by minimizing the denoising reconstruction error of the previous \name{} output $\h^k$,
\begin{equation}
\Wb^{k+1} = \argmin_{\Wb}\sum_{j=1}^m\sum_{i=1}^n\|\h^k_{i}-\Wb\tilde{\h}^k_{ij}\|^2.\label{eq:stacked}
\end{equation}
We solve eq.~(\ref{eq:stacked}) with the closed form solution for eq.~(\ref{eq:Wlimit}) as $m\!\rightarrow\! \infty$.  %analogously to eq.~(\ref{eq:ols}). 
%To obtain a stacked representation, Algorithm~\ref{code:matlab2} calls Algorithm~\ref{code:matlab} once for each layer. We refer to our algorithm as \snamelong{} (\sname{}). 
Figure~\ref{fig:flow} depicts a schematic layout of the work-flow of \sname{}. Algorithm~\ref{code:matlab2} shows a 9-lines MATLAB\tm implementation. 

\begin{algorithm}[t]
\begin{center}
\begin{Verbatim}[frame=none, framerule=0.1mm]
function [Ws,hs]=slide(X,p,t,l); 
    [d,n]=size(X);
    hs=X;
    Ws=zeros(d,d+1);
    for s=1:l 
     Ws(:,:,s)=lide(hs(:,:,s),p); 
     hst=double([hs(:,:,s)>t;ones(1,n)]);
     hs(:,:,s+1)=Ws(:,:,s)*hst;  
    end;
\end{Verbatim}
\end{center}
\caption{\sname{} in MATLAB\tm.\label{code:matlab2}}
\end{algorithm}

\section{SVM Training}
\label{sec:svm}
When the layers of \sname{} are computed, we regard them as features for SVM classification. For simplicity, we use the RBF kernel throughout this paper. 
%feed the values of these layers into a SVM with an RBF kernel by concatenating all the layers into one big vector. In our configuration, we have at most $4$ denoising layers stacked together, and thus for each input $\x_i$, including the raw input, we obtain at most $5$ layers of representations $\h^0_i,\dots,\h^4_i$. To simplify notation we refer to the raw input has $\h^0_i=\x_i$. 
The RBF kernel, described in~(\ref{eq:rbf}), accesses individual inputs only through pairwise distances. However, in each representation, the average distance between data points may vary significantly. Concatenating all the hidden layers and using a single kernel width $\sigma$ across all input features would over-emphasize the impact of some layers over others. We overcome this problem by introducing a specific kernel width for each layer $\sigma_0,\dots,\sigma_l>0$. The kernel function, for the case where  the first $l$ hidden layers (and the raw input data) are used, then becomes 
% \begin{equation}
% 	d(\x_i,\x_j)=([\frac{x_{1i}}{\sigma_1},\dots,\frac{x_{mi}}{\sigma_m}] - [\frac{x_{1j}}{\sigma_1},\dots,\frac{x_{mj}}{\sigma_m}])^\top([\frac{x_{1i}}{\sigma_1},\dots,\frac{x_{mi}}{\sigma_m}] - [\frac{x_{1j}}{\sigma_1},\dots,\frac{x_{mj}}{\sigma_m}])\label{eq:combinedkernel}
% \end{equation}
% where $\x_{1i} \in \textrm{layer 1}, \x_{mj} \in \textrm{layer m}$, and $\sigma_1$ is the average distance between all data points in layer $1$, and $\sigma_n$ is the average of layer $n$.
\begin{equation}
	k(\x_i,\x_j)=\exp{-\left(\frac{1}{\sigma^2}\displaystyle\sum_{t=0}^l\frac{\|\h^t_i-\h^t_j\|_2^2}{\sigma_t^2}\right)}.\label{eq:ksigma}
\end{equation}
%where $\h^{t}_i$ are the features from the $t^{th}$ layer generated by input $\x_i$, and $\sigma_t$ is the scaling factor for layer $t$ only. 
For our setup we 
%fix $\sigma_t$ to the median distance within the $t^{th}$ layer and choose the global $\sigma$ by cross validation, or 
set $\sigma=1$ and learn the individual $\sigma_t$ with the method described in the following subsection. 

%, \emph{i.e.} $\sigma_t^2=\frac{2}{n(n-1)}\sum_{i\neq j} \|\h_i^t-\h_j^t\|_2^2$.

%To further optimize the performance of the RBF kernel SVM, we use cross-validation to find a global scaler for the combined kernel, and a constant regularizer $C$ for the SVM. The combined kernel is,
% \begin{equation}
% 	d(\x_i,\x_j)=\frac{1}{\sigma^2}([\frac{x_{1i}}{\sigma_1},\dots,\frac{x_{mi}}{\sigma_m}] - [\frac{x_{1j}}{\sigma_1},\dots,\frac{x_{mj}}{\sigma_m}])^\top([\frac{x_{1i}}{\sigma_1},\dots,\frac{x_{mi}}{\sigma_m}] - [\frac{x_{1j}}{\sigma_1},\dots,\frac{x_{mj}}{\sigma_m}])\label{eq:combinedkernel2}
% \end{equation}
% \begin{equation}
% 	d(\x_i,\x_j)=\frac{1}{\sigma^2}\displaystyle\sum_{t=1}^m\frac{\|\h_{ti}-\h_{tj}\|}{\sigma_t^2}
% \end{equation}
%Since the cross-validation depends independently on each split, we can significantly increase this procedure by running it in parallel. 

% \begin{figure}[t]
% \centering
% \includegraphics[width=0.5\textwidth]{plots/rectangles.png}
% \includegraphics[width=0.5\textwidth]{plots/rectangles_images.png}
% \caption{The Rectangles dataset on the top, and the Rectangle-images dataset on the bottom. The task is to determine whether the rectangle has larger width or length.\label{figure:rectangles}} 
% \end{figure}
% 
% \begin{figure}[t]
% \centering
% \includegraphics[width=0.5\textwidth]{plots/convex.png}
% \caption{The Convex dataset. Each image contains only binary pixels (0=black and 1=white). The task is to determine whether an image contains a single white convex region.\label{figure:convex}} 
% \end{figure}

\textbf{Kernel Parameters Learning.} The features from the different layers $\h^0,\dots,\h^t$  might be of varying utility~\cite{Montavon2010} for the final discrimination task. 
%Also, as we show in the result section~\ref{sec_experiments}, on some data sets it is beneficial to include the original raw data $\h^0$, whereas in others it might actually decrease performance. 
We suspect that the exact values of $\sigma_t$ might have a noticeable influence on the classification accuracy. As the computation time of cross-validation grows exponentially with the number of parameters, we 
% The most common approach to SVM parameter setting is grid-search based on cross-validation. However, as we have more parameters than usual, the grid-search becomes impractical. 
%We did not try a grid search to set the parameters $\sigma_0,\dots,\sigma_t$ because of the exponential growth of combinations of candidate values with respect to the number of hidden layers $t$. 
%
investigate the benefits of learning the best values for $\sigma_0,\dots,\sigma_l$ automatically from the data. \cite{Chapelle02choosingmultiple} propose a straight-forward but effective algorithm to learn multiple kernel parameters for support vector machines with gradient descent. We use Chapelle's publicly available code\footnote{http://olivier.chapelle.cc/ams/} after a trivial modification\footnote{Each iteration we take a gradient-step with respect to all kernel widths instead of just the one global $\sigma$.} to learn all $\sigma_t$ in~(\ref{eq:ksigma}). 
We would like to emphasize that the whole time required for training the SVM \emph{and} optimizing for the individual kernel widths took on the order of minutes even for the larger data sets. As the code by~\cite{Chapelle02choosingmultiple} is only implemented for binary settings, % for the only multi-class problem (MNIST) we tuned the parameters on a binary sub-problem (\emph{threes} vs. \emph{eights}). 
we did not apply this technique to the multi-class dataset (MNIST). We also did not use kernel parameter learning for RBFs on raw input, as in this setting cross validation yielded better results. It is important to note that the kernel parameter learning equally applies to both SdA and \sname{}.

\section{Results}
\label{sec_experiments}
We evaluate our algorithm on several data sets from the deep-learning benchmark collection\footnote{http://tinyurl.com/64fgmzv \label{note:umontreal}}, including: The MNIST handwriting digits recognition dataset\footnote{http://yann.lecun.com/exdb/mnist/}, the Rectangles-images, the Rectangles and the Convex data set.

% \begin{figure}[htbp]
% \centering
% \includegraphics[width=0.45\textwidth]{plots/rectangles.png}
% \includegraphics[width=0.45\textwidth]{plots/rectangles_images.png}
% \caption{The Rectangles dataset on the top, and the Rectangle-images dataset on the bottom. The task is to determine whether the rectangle has larger width or length.\label{figure:rectangles}} 
% %\vspace{-2ex}
% \end{figure}
% 
% \begin{figure}[htbp]
% \centering
% \includegraphics[width=0.45\textwidth]{plots/convex.png}
% \caption{The Convex dataset. Each image contains only binary pixels (0=black, 1=white). The task is to determine whether an image contains a single white convex region.\label{figure:convex}} 
% \end{figure}

% \begin{table}[t]
%     \tabcolsep 3.8pt
%   	\small
% 	\vspace{0.25in}
% 	\center
% \begin{tabular}{|l|c|c|c|c|}
% 	\hline {\bf Statistics} &  {\bf MNIST} & {\bf Rect} & {\bf Rect-imgs} & {\bf Convex} \\
% 	\hline
% 	\hline \#features & 784 & 784 & 784 & 784 \\
% 	\hline \#training & 60000 & 1200 & 12000 & 8000 \\  
% 	\hline \#testing & 10000 & 50000 & 50000 & 50000 \\
% 	\hline \#folds & 1 & 50 & 10 & 10\\
% 	\hline
% \end{tabular}
% % \footnotetext[7]{due to time-constraints it was still missing at the deadline time}
% \caption{Statistics for all data sets. We averaged our results over 50 folds (same test set, but different train/validation splits) on Rect because the training set is significantly smaller than for the other data sets. MNIST was not averaged because of the abnormally large training set size.} 
% ~\label{table:datasets}
% \end{table}

\textbf{Datasets.}
The \emph{MNIST} data consists of 60,000 training and 10,000 testing images of handwritten digits $0-9$, where each image is of size $28\times 28$  pixels. The learning task is to predict the digit identity from the image. The \emph{Rectangles} dataset has $1,200$ training and $50,000$ testing images, also of size $28\times 28$. Each image contains a rectangle, whose border has a pixel value of $1$ (white), while all other pixels are $0$ (black). The learning task is to determine whether the rectangle has larger width or length. The \emph{Rectangles-Images} dataset is motivated by the same learning task, 
%is similar to the Rectangles dataset, 
except that the background and the rectangle are created of two noisy image patches. Also, it has $12,000$ training and $50,000$ testing images. 
%One fills the background, the other is in rectangular shape in the foreground. %The task is also to determine whether the rectangle has larger width or length. 
% Samples from these two rectangle datasets are shown in figure~\ref{figure:rectangles}.
The last dataset in our evaluation is the \emph{Convex} dataset. It consists of $8,000$ training and $50,000$ testing images, of size $28\times 28$. The images are made of binary pixels (where $0$ is black and $1$ white). The task is to determine whether all white pixels in a given image form  a single convex region. 
% Samples from the Convex datasets are shown in figure~\ref{figure:convex}. 
% All data statistics are summarized in table~\ref{table:datasets}.

\begin{figure}[t]
	\centering
		\includegraphics[width=0.65\textwidth]{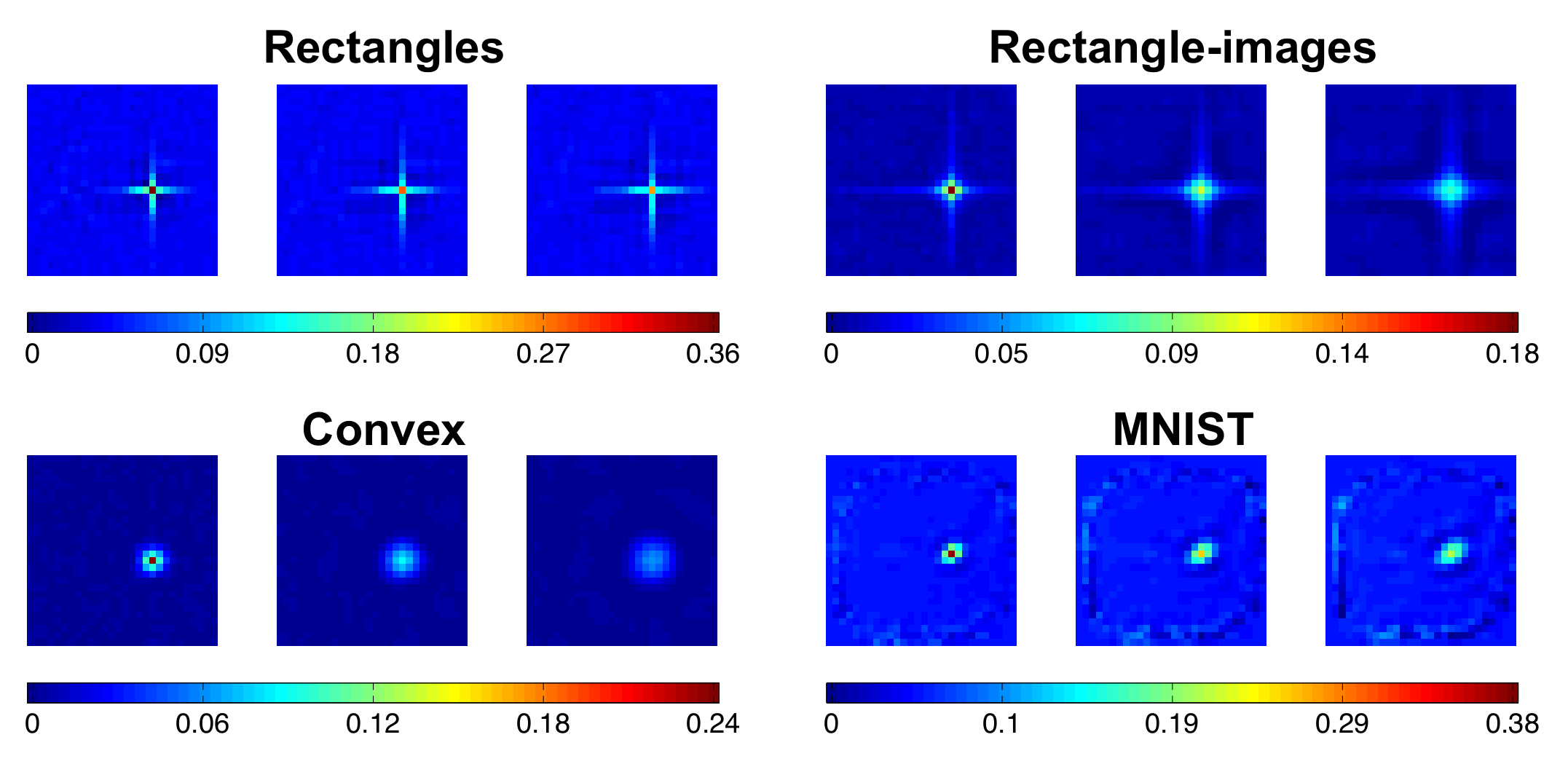}
	\caption{Reconstruction weights at different layers for the Rectangles, Rectangles-Images, Convex and MNIST data set. The reconstructed pixel itself is most correlated and therefore marked dark red. The cruciform lines in the top row illustrate that the de-noising weights have adapted to the shape of the rectangles. The convex and MNIST data set (bottom row) show a more spherical neighborhood structure, as no obvious neighborhood pattern is inherent in the data.}
	\label{fig:weights}                          
\end{figure}

% \begin{figure}[t]
% 	\centering
% 		\includegraphics[width=0.4\textwidth]{plots/dist_diff.pdf}
% 	\caption{Distance (in RBF kernel space) between these two same labeled images shrinks after denoising. Before denoising, these two images have only 2 pixels in common. After blurring, they get much closer than before, which significantly increases the chance of being classified correctly.}
% 	\label{fig:distance}                          
% \end{figure}

\textbf{Weight Matrices.}
Figure~\ref{fig:weights} visualizes the  reconstruction weights for four example pixels on Rectangles, Rectangles-Images, Convex and the MNIST data set at various layers of \sname{}. Each image shows one row of $\Wb^k$ (without bias weight), where the color reflects the weight value. The figure shows a clear trend towards ``fuzzier'' reconstruction as the depth increases.  In both rectangles data sets (top row) each pixel is reconstructed from neighboring pixels with a tendency towards vertical \emph{or} horizontal offset, thus incorporating the inherent structure from the rectangles in the data. The clarity of this trend is particularly impressive for the Rectangles-Images data set, where the rectangles consist of very noisy image patches. In the convex and MNIST data sets, the pixel is reconstructed from a small circular patch of surrounding pixels. In the case of MNIST, small non-zero weights also exist in distant corner locations, originating from the fact that some pixels may only be non-zero in one or two images in the training data set.

\textbf{Experiments Settings.}
For all data sets, the training and testing splits are pre-defined. We create an additional random $80/20$ split on the training set for the purpose of parameter tuning. %Table~\ref{table:datasets} lists the important data set statistics.

To find the best combination of noise level $p$ and de-noising layers $\ell$, we run cross-validation on the validation set. As the features across all data sets are naturally within the interval $[0,1]$, we set the threshold parameter to $t=0.5$ for all experiments.\footnote{This choice can be further justified as the squared loss predictor approximates the probability that a binary target has value $1$~\cite{Bishop06} making $0.5$ the ideal cutoff to minimize the prediction error.} 
%The parameters of RBF-kernel SVM, kernel width $\sigma$ and constant $C$, are set automatically by kernel parameter learning if the data set has binary labels, or set by cross validation otherwise. 
%$training parameters
As described in the previous section, we used kernel parameter learning and SVMs with a RBF kernel~\cite{cortes1995support} for our classification, which we refer to as (\sname{}-SVM).  
For comparison, we evaluate against three other methods:  We use SVMs with a RBF kernel on the raw input (\emph{Raw-SVM}), where we use cross-validation to select the kernel width $\sigma^2$ and the regularization trade-off $C$, as mentioned in the previous section. 
%\footnote{Using automated kernel parameter setting for just one kernel-width resulted in slightly worse accuracy results than cross-validation. As this is our baseline and not the focus of this paper, we do not report these results here.} 
Further, we use deep neural networks (\emph{SdA}), which were pre-trained with SdA~\cite{Vincent08extractingand} and fine-tuned with back-propagation. We use the network architecture, noise level, and learning rates recommended by the authors of~\cite{Larochelle07anempirical} through personal communication. 
We also use the hidden representations of the SdA as input for the SVM, (\emph{SdA-SVM}), in the exact same fashion as \sname{}-SVM. Finally, other linear transformation are also evaluated, such as PCA, Whitening and random projection (\emph{PCA-SVM, White-SVM, Rand Proj-SVM}). For PCA and Whitening, we reduce the feature dimension by only keeping $p$ percent of variance, and the $p$ is selected by cross-validation.  
%Similar to other SVM based classification, we set the $\sigma$ and $C$ by cross-validation on the validation set. 
%The second setup is having the neural network fine-tuned with back-propagation and early-stopping based on the validation data(tuning rate and stop criteria are following the setup of~\cite{Vincent08extractingand}).

%For RAW-SVM, we feed the SVM with the original input of the data sets and . For neural networks, we have two setups. Both are pre-trained layer by layer as DAs, . 
%The difference is at the classification stage. 

\begin{figure}[t]
	\centering
		\includegraphics[width=0.65\textwidth]{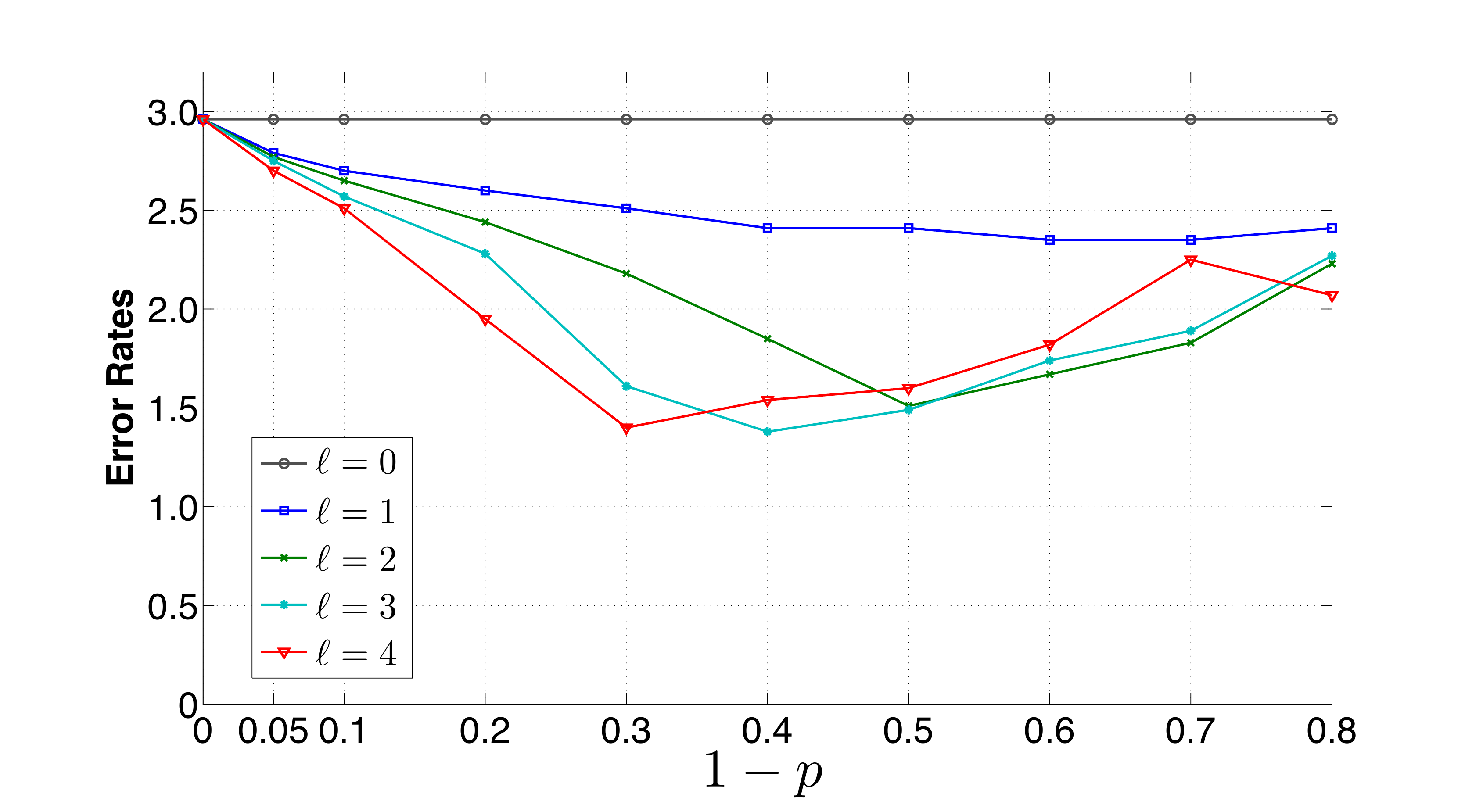}
	\caption{The Rectangles test error as a function of the noise level $(1\!-\!p)$ and number of layers $\ell$. The graph shows clearly that deep layers provide a noticeable improvement up to $\ell=2$ layers.}
	\label{fig:trend}                          
\end{figure}

\textbf{Parameter sensitivity.}
Figure~\ref{fig:trend} displays the classification error on the Rectangles data set as a function of the noise level $p$. The four colored lines correspond to different depths $\ell=0,\dots,4$, where $\ell=0$ or $p=1$ correspond the original raw input\footnote{The results for $p=1$ do not match table~\ref{table:results} because for this graph the kernel parameters were learned, which is sub-optimal for scenarios with few parameters.} not processed with \sname{}. In general, \sname{} appears to be somewhat sensitive to the exact choice of $p$ and $\ell$, and we choose it by cross validation for all experiments. There is a clear trend that deep layers $\ell>1$ improve over a single layered transformation $\ell=1$.

\begin{table}[t]
    \tabcolsep 3.8pt
  	\small
	\vspace{0.25in}
	\center
\begin{tabular}{|l|c|c|c|c|c|c|c|}
	\hline 
	{\bf Error in \%} &  Raw-SVM & SdA & SdA-SVM & \sname{}-SVM & Rand Proj-SVM & PCA-SVM & White-SVM \\
	\hline {\bf MNIST} & 	1.49  & \textbf{1.31}  & 1.41   & 1.36 	& 1.47 & 1.40 & 1.59 \\
	\hline {\bf Rect}  &	2.45	& 2.43 	& 1.10	& 1.38  & 2.94 & 1.54 & \textbf{0.96} \\
	\hline {\bf Rect-imgs}	& 23.29	& 23.07 	& 22.70	& \textbf{22.39} & 23.27 & 22.55 & 22.58 \\
	\hline {\bf Convex} &	18.72	& 17.51		& 17.59	& 12.18 & 18.13 & \textbf{11.93} & 18.41 \\
	% \hline {\bf Cifar} &	56.06	&  	&  	& \textbf{55.04} & 55.67 & 56.30 & 56.31 \\
	\hline
\end{tabular}
% \footnotetext[7]{due to time-constraints it was still missing at the deadline time}
\caption{Error rates (in percent) of deep neural networks with stacked auto-encoder (SdA) pre-training, Support Vector Machines without pre-training (Raw-SVM) and SVMs with \sname{} features (\sname{} SVM).} 
~\label{table:results}
\end{table}

% \begin{table}[t]
% 	\footnotesize
% 	\begin{center}
% 	\begin{tabular}{|l|c|c|c|}
% 		\hline
% \textbf{Error in \%}	& \textbf{Rect} & \textbf{Rect-Images} & \textbf{Convex} \\
% \hline
% 	\hline
% 	SdA-SVM & 1.10 & 22.70 & 17.59 \\
% 	\hline 
% 	\sname{}-SVM & 1.39 & 22.24 & 12.21 \\	
% 	\hline
% \end{tabular}
% \end{center}
% \caption{Direct comparison between the features of SdA and \sname{} on all data sets with binary labels. The SVM parameters were set with the gradient-descent method by~\cite{Chapelle02choosingmultiple}.}
% \end{table}

% \begin{table*}
% 	\tabcolsep 3.8pt
%   	\small
% 	\vspace{0.25in}
% 	\center
% \begin{tabular}{|l|l|}
% 	\hline {\bf Algorithm:} & {\bf Parameters} \\
% 	\hline 
% 	\hline SdA & $\bullet$layers $\bullet$noise $\bullet$hidden units $\bullet$pre-training epochs $\bullet$pre-training rate $\bullet$fine-tuning rate\\		
% 	\hline SdA SVM& $\bullet$layers $\bullet$noise $\bullet$hidden units $\bullet$pre-training epochs $\bullet$pre-training rate $\bullet$SVM $\sigma$ $\bullet$SVM $C$\\
% 	\hline Raw SVM &	$\bullet$SVM $\sigma$ $\bullet$SVM $C$\\
% 	\hline CFNN SVM & $\bullet$layers $\bullet$noise $\bullet$SVM $\sigma$ $\bullet$SVM $C$\\
% 	\hline
% \end{tabular}
% \caption{Parameters. This table shows parameters required by different algorithms. Raw data SVM has the least amount of parameters, and Convex Denoising has only 2 more parameters, which can be found by cross-validation on a small validation set.} 
% ~\label{table:parameters}	
% \end{table*}      

\textbf{Experiments Results.}
All classification results are shown in table~\ref{table:results}. A few general trends can be observed. First, using \sname{} feature pre-processing yields  considerable improvements over results with original features (Raw-SVM) on \emph{all} data sets. In fact, transforming the features with \sname{} makes SVM outperform even Deep Neural Nets (SdA) in all but one (MNIST) learning tasks. Finally \sname{} even outperforms the much more complex SdA features on two data sets (Rect-imgs,Convex) and obtains equivalent results (up to significance) on MNIST. 
%When SdA is used to generate features for SVM  (SdA-SVM) it obtains significantly lower error rates than \sname{} only on the Rect data set. 
Especially on Convex and the Rect-Images data sets, \sname{} clearly outperforms all other algorithms. 
Results from PCA (PCA-SVM) are also significantly better than original features (Raw-SVM), and is the best in the Convex dataset, but worse than SLIDE on all others. Whitening (White-SVM) tops the Rectangle dataset, but does not yield significant improvements on other datasets.
These results are very encouraging as \sname{} is trivial to implement and orders of magnitudes faster than SdA. 

% has the best results on 3 out of 4 data sets, and outperforms SVM fed by raw data on all 4 data sets, with very little overhead. Neural Network with fine-tuning has the best result on MNIST, and also slightly better than neural network with SVM on the Convex data set. Table~\ref{table:parameters} compares the parameters needed to be settled for each method. Raw data SVM has only two parameters SVM $\sigma$ and $C$. Convex Denoising has two more parameters, number of layers and the level of noise, which can be found easily by running cross-validation on a small validation set. If we assume that all the parameters are given, Convex Denoising is still very fast, only a little slower than the regular SVM. Since the solution to the de-noising matrix is closed-form, we don't need to do any gradient descent, but just some matrices multiplication, which is really fast in Matlab. The comparison of running time is shown in table~\ref{table:runningtime}.

\textbf{Running Time.}  Table~\ref{table:runningtime} compares the running times for feature generation with  SdA and \sname{}. 
All timings are performed on a desktop with dual Intel\tm  Six Core Xeon X5650 2.66GHz processors. 
To train the SdA we use the highly optimized Theano open-source package\footnote{http://deeplearning.net/software/theano/} which is carefully parallelized and in our experiments utilizes a state-of-the-art GPU\footnote{NVIDIA Quadro FX 1800 768MB GDDR3} with additional 64 cores. No explicit parallelization was used for \sname{}. The results show a three orders of magnitude speed-up across all data sets, reducing the pre-training time from several hours to a few seconds. 

\begin{table}[h]
    \tabcolsep 3.8pt
  	\small
	\vspace{0.25in}
	\center	
\begin{tabular}{|l|c|c|c|c|}
	\hline {\bf Time} &  {\bf MNIST} & {\bf Rect} & {\bf Rect-imgs} & {\bf Convex}\\
	\hline 	
	\hline SdA & 8h  24m & 1h 30m & 2h 40m & 6h 50m \\		
	\hline \sname{} & 22s & 5s & 8s & 6s \\
	\hline
	\hline $\times$ speedup & 1377 & 1080 & 1207 & 4100 \\
	\hline
\end{tabular}
\caption{Running time required for feature generation with SdA and \sname{}.} ~\label{table:runningtime}
\end{table}

\section{Related Work}
\label{sec_related}

Related work can be categorized into three lightly correlated dimensions: layer-wised unsupervised pre-training, learning from partially corrupted data, and linking SVMs with neural networks. In the area of layer-wised unsupervised pre-training, \cite{Hinton06fast}, \cite{lee2009unsupervised}, \cite{Bengio07greedylayer-wise}, \cite{lee2007sparse} and \cite{marc2007sparse} provide various successful approaches. They all propose to pre-train the neural network by some unsupervised training criterion as an initialization for back-propagation or as features for other algorithms (e.g. SVM). In contrast, our method does not require any training through gradient descent and is orders of magnitudes faster. 
%Different from their approach, we pre-train representations not exclusively for neural networks, and can be fed to any other classifiers. 
% Multiple publications have investigated learning presentations from partial corrupted data. 

%\cite{lee2009unsupervised} investigate using the intermediate representations from neural networks for SVM on audio data and achieve significantly better classification results, but their approach is different from us as they still use neural networks to do the pre-training. 

\cite{Rahimi08} use carefully constructed random projections to create features that approximate kernelization for linear SVMs. In the area of learning from partially corrupted data, \cite{Vincent08extractingand} proposes using Stacked Denoising Auto-encoder(SdA)  to reconstruct the corrupted data, and demonstrates that the learnt representations are robuster. Our method is heavily inspired by their work and can be viewed as a convex closed-form transformation that mimics their feature generator. 

\cite{cho2010large} also investigate linking neural networks with SVMs through deep kernels. In their work, they construct new recursively composed positive semi-definite kernel functions which can be viewed as mimicking the layer by layer training of neural networks. Different from our work, their method is not based on feature de-noising and still requires computationally expensive ``fine-tuning'' through distance metric learning~\cite{weinberger2009distance}.

\section{Conclusion}
\label{sec_conclusion}
%Future work: Other classifiers than SVM
%
%
%
We introduced \sname{}, a novel algorithm based on stacked linear denoising autoencoders for extremely fast layer-wise deep pre-training for feature generation. We derived a simple closed-form solution that can be implemented in a few lines of MATLAB\tm. 
We demonstrated that \sname{} for SVM classification can match (or out-perform) the classification results of SdA features -- on three out of four data set even beyond the low error rates of deep neural networks. Most notably, the running time of \sname{} is reliably in the order of seconds even on data sets with tens of thousands of data points. 

As future directions we plan to investigate other classifiers and different learning settings. As \sname{} is entirely unsupervised, it lends itself naturally towards semi-supervised and transfer learning tasks. 

Because \sname{} is so straightforward to implement, only takes seconds to compute and improves results for SVM classification with surprising consistency, we have great hopes that it will find use as a general pre-processing algorithm across many areas of machine learning.

% \textbf{Acknowledgements}\\
% \noindent
% We thank the creators of Theano, and the creators of the deep-learning benchmark data sets without whom this research would not have been possible. 

%\section{Acknowledgements}
%\label{sec_acknowledgements}
%\input acknowledgements.tex

% %\small
\bibliographystyle{abbrv}
\bibliography{deepnets}

\end{document}